\documentclass[11pt]{article}

\usepackage[final]{acl}

\usepackage{times}
\usepackage{latexsym}
\usepackage[T1]{fontenc}

\usepackage[utf8]{inputenc}
\usepackage{enumitem}

\usepackage{microtype}
\usepackage{float}
\usepackage{inconsolata}

\usepackage{graphicx}
\usepackage{amssymb}
\usepackage{amsmath}
\usepackage{subcaption}
\usepackage{booktabs}
\usepackage{multirow}
\usepackage[linesnumbered,ruled,vlined]{algorithm2e}

\title{LocalSUG: City-Preference-Enhanced LLM for Query Suggestion in Local-Life Services}

\author{
 \textbf{Jinwen Chen}\textsuperscript{1,2}\thanks{Equal contribution.},
 \textbf{Shiwen Zhang}\textsuperscript{1,2}\textsuperscript{*},
 \textbf{Shuai Gong}\textsuperscript{2},
 \textbf{Zheng Zhang}\textsuperscript{2},
 \textbf{Yachao Zhao}\textsuperscript{2},
 \\
 \textbf{Lingxiang Wang}\textsuperscript{1},
 \textbf{Haibo Zhou}\textsuperscript{2},
 \textbf{Wei Lin}\textsuperscript{2},
 \textbf{Hainan Zhang}\textsuperscript{1}\thanks{Corresponding author: zhanghainan1990@163.com}
\\
\textsuperscript{1}Beijing Advanced Innovation Center for Future Blockchain and Privacy Computing
\\
 \textsuperscript{2}Meituan
 \\
}

\begin{document}
\maketitle
\begin{abstract}

In local-life service platforms, query suggestion reduces user effort by generating candidate queries from input prefixes. Traditional multi-stage systems rely heavily on historical popular queries, limiting their ability to capture long-tail and emerging demand. Although LLMs provide strong semantic generalization, their deployment in local-life services faces three challenges: insufficient city-preference awareness, exposure bias in preference optimization, and strict online latency constraints.
We propose LocalSUG, an LLM-based query suggestion framework for local-life services. LocalSUG mines city-preference-enhanced candidates from term co-occurrence and injects them into prompts as dynamic references rather than fusing them into model parameters. This allows the model to adapt to changing city preferences, such as merchant openings or closures, while reducing stale or locally invalid suggestions. We further introduce a beam-search-driven GRPO algorithm to align training with inference-time decoding and optimize relevance together with business-oriented rewards. Finally, quality-aware beam acceleration and vocabulary pruning reduce online latency while preserving generation quality. Offline evaluations and large-scale online A/B testing show that LocalSUG improves CTR by +0.35\% and reduces the no-result rate by 3.98\%, demonstrating its effectiveness in real-world deployment.
\end{abstract}

\section{Introduction}

%







\begin{figure}
    \centering
    \includegraphics[width=\linewidth]{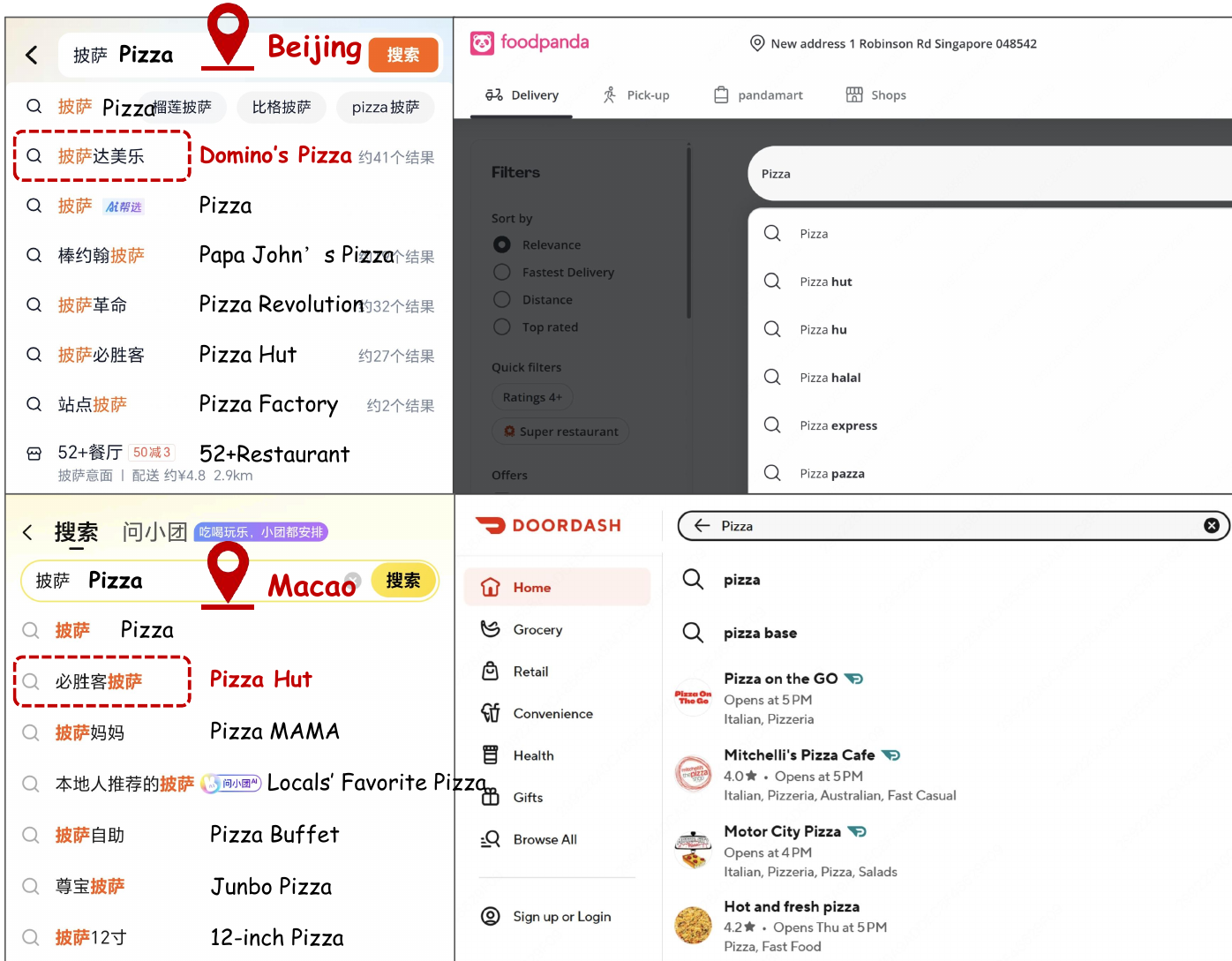}
    \caption{Examples of query suggestion in local-life service platforms.}
    \label{fig:app}
\end{figure}

Query suggestion~\cite{QS} is a critical component of search systems in local-life service platforms~\cite{LocalLifeServices,LocalSearchBench,FIM}, where users seek city-dependent services such as restaurants, hotels, entertainment, and home services, as shown in Figure~\ref{fig:app}. By predicting candidate queries from user-typed prefixes, query suggestion reduces typing effort, accelerates search, and guides users toward high-quality intents.

Most production systems~\cite{AutoFAS, taobao, YouTube, RankFlow} adopt a Multi-stage Cascading Architecture following a recall–ranking paradigm. While robust and scalable, such systems rely heavily on historical query logs and fixed candidate pools. As a result, they struggle to explore long-tail, compositional, or emerging intents that are insufficiently represented in past data. Recent generative approaches, such as OneSug~\cite{guo_onesug_2025}, replace retrieval with large language models (LLMs) that directly synthesize suggestions. Owing to their strong semantic generalization ability, LLMs can generate coherent and previously unseen suggestions, offering a promising solution to long-tail sparsity.

However, directly deploying LLMs for query suggestion in local-life platforms introduces three key challenges. \textbf{1) Insufficient city-preference awareness}: User intents differ across cities due to merchant availability, regional trends, and local habits. As shown in Figure~\ref{fig:app},  ``pizza'' may suggest Domino's in Beijing but be invalid in cities without available stores. As city preferences change with merchant openings and closures, LLMs without city-specific references may produce locally invalid suggestions, degrading user experience and business performance. \textbf{2) Exposure bias in preference optimization}: Existing generative methods use preference optimization (e.g., S-DPO~\cite{s-dpo}) trained at the sequence level on historical logs, but real-world systems generate and rank suggestions list-wise via beam search, creating a mismatch that can harm coherence and ranking consistency. \textbf{3) Online inference latency}: Local-life search systems operate under strict real-time latency requirements. High-capacity LLMs provide strong generative quality but incur substantial computational overhead, making naive deployment infeasible for large-scale production environments.

To address these challenges, we propose LocalSUG, an end-to-end generative query suggestion framework tailored for local-life service platforms. First, we design a city-preference-enhanced candidate mining strategy based on term co-occurrence. Instead of directly encoding city preferences into model parameters, LocalSUG injects daily updated city-specific candidates into the prompt as external references, allowing the model to follow current city preferences while avoiding stale memorization. Second, we introduce a Beam-Search-Driven GRPO algorithm that aligns training with inference-time decoding and optimizes sequence-level objectives through a multi-objective reward mechanism balancing relevance and business metrics. Finally, we develop Quality-Aware Accelerated Beam Search (QA-BS) and vocabulary pruning techniques to substantially reduce online latency while preserving generation quality, enabling the large-scale deployment of a 0.6B-parameter LLM in high-concurrency industrial environments.

Extensive experiments on real-world logs show that LocalSUG consistently outperforms strong baselines on HitRate@K, MRR, Diversity, and Quality, while the proposed acceleration strategy achieves a 3.11$\times$ decoding speedup. In online A/B tests, LocalSUG reduces the no-result rate by 3.98\%, boosts PV CTR by 0.35\%, and increases Unique Item Exposure by 7.14\%, demonstrating its effectiveness in real-world deployment.

The main contributions are as follows:
\begin{itemize}
    \item We propose a production-ready city-preference-enhanced query suggestion framework tailored for local-life services, enabling reliable generation under diverse and evolving city preferences.
    \item  We introduce Beam-Search-Driven GRPO to bridge offline training and real-time list-wise inference, improving ranking stability, diversity, and consistency in production settings.
    \item We introduce QA-BS and vocabulary pruning techniques to overcome latency bottlenecks, enabling the large-scale deployment of a 0.6B-parameter LLM in high-concurrency industrial environments.
    \item Extensive offline experiments and online A/B tests verify that LocalSUG significantly reduces no-result cases and improves CTR, demonstrating its superior effectiveness for query suggestion.
\end{itemize}

\begin{figure*}
    \centering
    \includegraphics[width=\textwidth]{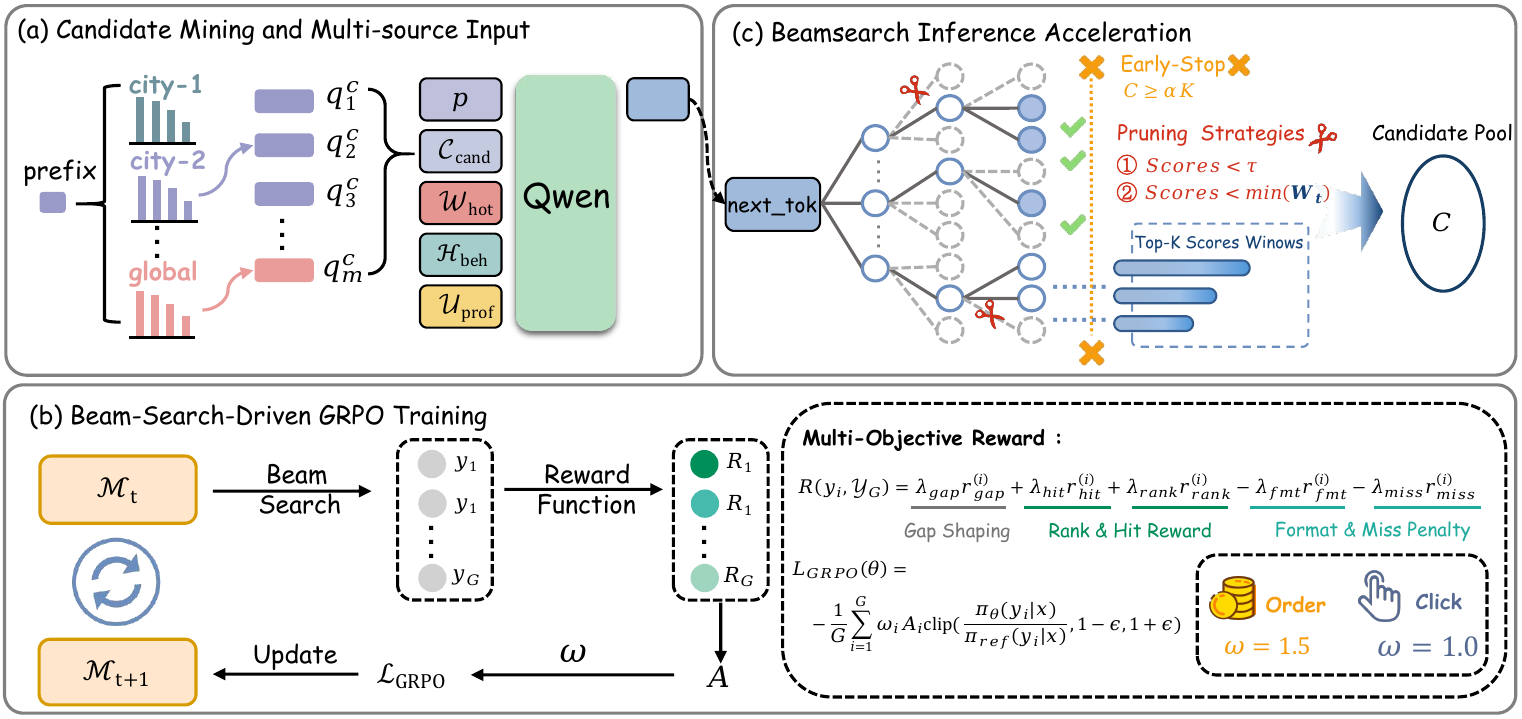}
    \caption{Overview of our proposed generative Query Suggestion framework. 
    (a) \textbf{Candidate Mining and Multi-Source Input}: Prefix-to-query candidates are retrieved via city-preference-enhanced and global co-occurrence rules, then concatenated with hot words, user history, and profiles. 
    (b) \textbf{Beam-Search-Driven GRPO Training}: The model is optimized using group relative rewards to bridge the gap between training and inference, incorporating hit, rank, and format objectives.
    (c)  \textbf{Beam Search Inference Acceleration}: Online performance is enhanced through Quality-Aware Accelerated Beam Search and LLM head pruning.}
    \label{fig:main}
\end{figure*}

\section{Methodology}

\subsection{System Overview}
We propose a unified generative framework for local-life services (e.g., food delivery) using \textbf{Qwen3-0.6B}~\cite{qwen3technicalreport} as the backbone to balance contextual intelligence with strict latency requirements. The overall architecture is illustrated in Figure~\ref{fig:main}, which consists of three core phases: (a) candidate mining and multi-source context construction; (b) Beam-Search-Driven GRPO training for multi-objective optimization; and (c) quality-aware inference acceleration for online deployment. The model $\mathcal{M}$ processes a concatenated input $\mathcal{X}$:
\begin{equation}
\mathcal{X} = \text{Concat}\big(p, \mathcal{C}_{\text{cand}}, \mathcal{W}_{\text{hot}}, \mathcal{H}_{\text{beh}}, \mathcal{U}_{\text{prof}}\big)
\end{equation}
consisting of: 1) user prefix $p$; 2) candidate queries $\mathcal{C}_{cand}=\{q^c_1, \dots, q^c_m\}$ fetched from a city-preference-enhanced co-occurrence cache; 3) trending hot words $\mathcal{W}_{hot}$ from internal platform services; 4) behavior history $\mathcal{H}_{beh}$; and 5) user profile $\mathcal{U}_{prof}$. The model generates a ranked list $\mathcal{Q} = [q_1, \dots, q_K]$ via beam search with width $K$. A seed model $\mathcal{M}_t$ is obtained through standard SFT.

\subsection{Prefix-to-Query Candidate Mining}

City preferences in local-life services evolve continuously with merchant openings and closures, seasonal trends, and regional consumption shifts. Directly fusing such preferences into model parameters would require frequent retraining and may cause stale memorization. For instance, if all stores of a chain brand close in a city, a parameterized model may still generate that brand as a locally plausible suggestion. We therefore treat city preferences as dynamic external references and inject them into the prompt through candidate queries.

Specifically, we adopt a statistical co-occurrence strategy rather than RQ-VAE~\cite{rqvae}, which introduces online inference overhead and is sensitive to vocabulary distribution shift. We directly extract prefix-query co-occurrence from historical logs. For each prefix $p$, we maintain two categories of sorted candidate lists based on click counts: \textbf{city-specific lists} $\mathcal{L}_p^{\text{city}}$ for each city and a \textbf{global list} $\mathcal{L}_p^{\text{global}}$. During prompt construction, we prioritize $\mathcal{L}_p^{\text{city}}$ to reflect current city preferences; if city-specific candidates are insufficient for $m$ slots, we backfill using top queries from $\mathcal{L}_p^{\text{global}}$. This design preserves local precision while ensuring coverage robustness. To capture temporal changes, the co-occurrence candidate pool is updated daily using a sliding window of the past seven days' logs.

\subsection{Beam-Search-Driven GRPO}
To bridge the training-inference gap, we employ Beam-Search-Driven Group Relative Policy Optimization (GRPO). Let $\pi_\theta$ denote the policy model and $\pi_{ref}$ denote the reference model. For each input $x$, we sample a group of $G$ outputs $\mathcal{Y}_G = \{y_1, \dots, y_G\}$ based on beam search from the policy $\pi_{\theta}$, where $G>K$. The modified GRPO loss is:
\begin{equation}
\begin{aligned}
&\mathcal{L}_{GRPO}(\theta) = \\ &- \frac{1}{G} \sum_{i=1}^{G} \omega_i \cdot \mathbb{A}_i \cdot 
\text{clip}\left( \frac{\pi_\theta(y_i|x)}{\pi_{ref}(y_i|x)}, 1-\epsilon, 1+\epsilon \right)
\end{aligned}
\end{equation}
where $\mathbb{A}_i = (R(y_i, \mathcal{Y}_G) - \bar{R})/(\sigma_R + \delta)$ and $\omega_i = \lambda_{order}$ if $x$ leads to a conversion (else $1.0$). The unified reward function $R(y_i, \mathcal{Y}_G)$ combines five objectives:
\begin{equation}
\begin{split}
R(y_i, \mathcal{Y}_G) =  \underbrace{\lambda_{gap} \cdot r_{gap}^{(i)}}_{\text{Gap Shaping}} + &\underbrace{\lambda_{hit} \cdot r_{hit}^{(i)}}_{\text{Target Hit}} \\
+ \underbrace{\lambda_{rank} \cdot r_{rank}^{(i)}}_{\text{Rank Bonus}} - \underbrace{\lambda_{fmt} \cdot r_{fmt}^{(i)}}_{\text{Format Penalty}} -& \underbrace{\lambda_{miss} \cdot r_{miss}^{(i)}}_{\text{Miss Penalty}}.
\end{split}
\end{equation}
Key mechanisms: 
1) \textbf{Gap Shaping}: We penalize the tail $G-K$ candidates to force high-quality results into the top-$K$ beam. 
2) \textbf{Rank \& Hit}: If ground truth $y^*$ exists in $\mathcal{Y}_G$, we assign a hit bonus and a rank reward $1/\log_{10}(rank(y^*) + 1)$, while penalizing non-target candidates ranked higher than $y^*$. 
3) \textbf{Format \& Miss}: If $y^*$ is missing or format errors (repetition/invalid tokens) occur in the top-$K$, we penalize the errors and boost valid candidates from the tail to fill the $K$ slots.
The detailed calculation process is described in Appendix~\ref{app:reward_calculation}.

\begin{table*}[!t]
  \centering
  \caption{\label{tab:main}Offline performance comparison between LocalSUG and various baselines. The best results are highlighted in \textbf{bold}, and the second-best results are \underline{underlined}. For more comparisons across different LocalSUG variants, please refer to Appendix~\ref{app:main2}.}
  \resizebox{\textwidth}{!}{
    \begin{tabular}{l|cccc|cccc}
    
    \toprule
    \multirow{2}{*}{Method} & \multicolumn{4}{c|}{\textsc{Click}} & \multicolumn{4}{c}{\textsc{Order}} \\
     & HR@12↑ & MRR↑ & DIV↑ & QUA↑ & HR@12↑ & MRR↑ & DIV↑ & QUA↑ \\

    \midrule
    MCA & 74.24\% & 40.62\% & 62.85\% & \underline{99.94\%} & 76.64\% & 42.78\% & 63.80\% & \underline{99.93\%} \\
    City-MCA & 74.43\% & 45.93\% & 61.00\% & \textbf{99.95\%} & 75.77\% & 47.87\% & 61.92\% & \textbf{99.96\%} \\
    LMP & 79.67\% & 40.81\% & 53.57\% & 99.75\% & 79.66\% & 39.99\% & 67.81\% & 99.84\% \\
    SPQR & 95.95\% & 70.53\% & 45.49\% & 99.83\% & 96.28\% & 70.84\% & 63.75\% & 99.85\% \\
    SFT & 96.07\% & 79.36\% & 63.67\% & 90.43\% & 96.63\% & 79.39\% & 61.91\% & 90.80\%  \\
    OneSug & 73.19\% & 54.83\% & \textbf{72.18\%} & 99.50\% & 77.67\% & 59.26\% & \textbf{70.28\%} & 99.52\% \\
    $\text{OneSug}_{rule}$ & \textbf{96.47\%} & \underline{80.69\%} & 68.62\% & 96.17\% & \underline{97.07\%} & \underline{81.25\%} & 66.61\% & 96.54\% \\
    LocalSUG & \underline{96.36\%} & \textbf{81.13\%} & \underline{71.49\%} & 98.55\% & \textbf{97.09\%} & \textbf{81.80\%} & \underline{69.31\%} & 98.78\% \\ 
      
    \bottomrule
    \end{tabular}%
    }

\end{table*}%

\subsection{Beam Search Inference Acceleration}

To facilitate online deployment, we propose \textbf{Quality-Aware Accelerated Beam Search} (QA-BS, see Algorithm~\ref{alg:qa_beam_search} for the detailed inference logic):
\begin{itemize}
    \item \textbf{Quality Verification}: Active beams are pruned if cumulative log-probability $S(y) < \tau$. Completed sequences are accepted only if $S(y) > \min(W_t)$, where $W_t$ are the current top-$K$ scores.
    \item \textbf{Adaptive Termination}: Early exits are triggered if valid candidates saturate capacity ($|C| \ge \alpha K$) or if all active beams fall below $\tau$ (Fail-Safe), preventing high-latency tail processing.
\end{itemize}
Furthermore, statistical analysis of 38 million log entries reveals sparse token frequency. We pruned the LM head to the top 30,000 most frequent tokens, significantly reducing computation for irrelevant tokens with minimal impact on accuracy.

\begin{table*}[!t]
  \centering
    \caption{\label{tab:abl}Ablation Study of LocalSUG in Offline Evaluations. The best results are in bold.}
  \resizebox{\textwidth}{!}{
    \begin{tabular}{l|cccc|cccc}
    
    \toprule
    \multirow{2}{*}{Method} & \multicolumn{4}{c|}{\textsc{Click}} & \multicolumn{4}{c}{\textsc{Order}} \\
     & HR@12↑ & MRR↑ & DIV↑ & QUA↑ & HR@12↑ & MRR↑ & DIV↑ & QUA↑ \\

    \midrule
    LocalSUG & 96.36\% & \textbf{81.13\%} & 71.49\% & 98.55\% & \textbf{97.09\%} & \textbf{81.80\%} & 69.31\% & \textbf{98.78\%} \\

    \midrule
    - w/o $r_{gap}$ & 95.20\% & 80.97\% & \textbf{73.31\%} & 97.37\% & 96.02\% & 81.66\% & \textbf{72.58\%} & 97.71\% \\

    \midrule
    - w/o $r_{hit}$ & 95.88\% & 81.09\% & 72.45\% & 97.78\% & 96.74\% & 81.74\% & 71.44\% & 98.45\% \\
    
    \midrule
    - w/o $r_{rank}$ & \textbf{96.58\%} & 80.00\% & 70.73\% & \textbf{98.99\%} & 97.07\% & 80.59\% & 68.19\% & 96.83\% \\

    \midrule
    - w/o $r_{fmt}$ & 96.03\% & 79.32\% & 63.62\% & 90.36\% & 96.74\% & 77.59\% & 62.91\% & 89.85\% \\

    \midrule
    - w/o $r_{miss}$ & 95.62\% & 81.04\% & 72.73\% & 97.49\% & 96.38\% & 81.71\% & 72.02\% & 97.82\% \\
    
    \midrule
    - w/o $\omega$ & 96.32\% & 81.11\% & 71.62\% & 98.50\% & 96.72\% & 80.63\% & 71.55\% & 98.24\% \\
    
    \bottomrule
    \end{tabular}%
    }

\end{table*}%

\section{Experiments}

\subsection{Experimental Settings}

\paragraph{Datasets} We collected 8 consecutive days of real-world exposure and click logs from our local-life service platform. The first 7 days served as the training set (3.5M samples), and the $8^{th}$ day as the test set (50k samples). For evaluation, we constructed three subsets from the test data: 1) \textsc{Mix}: 10,000 randomly sampled instances; 2) \textsc{Click}: 10,000 instances with user clicks; and 3) \textsc{Order}: 5222 instances leading to a transaction.

\paragraph{Metrics} We evaluate performance using \textbf{HitRate@K} (HR@K) and \textbf{MRR} (Mean Reciprocal Rank). Additionally, we report \textbf{Diversity (DIV)}, defined as the ratio of unique queries to total generated queries, and \textbf{Quality (QUA)}, measuring the proportion of generated queries that are free from formatting errors (e.g., garbled text, invalid tokens) and, crucially, contain no \textit{intra-list duplicates} within the top-$K$ outputs for a single input. See Appendix~\ref{sec:appendix_metrics} for further details.

\paragraph{Baselines}
We compare LocalSUG with four groups of baselines: 
(1) industrial cascading systems, including MCA and its city-enhanced variant City-MCA; 
(2) location-aware search baselines adapted from web search, including LMP~\cite{lmp} and SPQR~\cite{spqr}; 
(3) generative query suggestion baselines, including OneSug~\cite{guo_onesug_2025} and OneSug$_{rule}$; and 
(4) LocalSUG variants, including DPO-based training~\cite{dpo}, a Llama backbone, and different GRPO sampling strategies. 
Detailed descriptions of all baselines are provided in Appendix~\ref{app:baseline}.

\paragraph{Implementation Details} Experiments were conducted on H20-141G.
We utilized \textbf{Qwen3-0.6B}~\cite{qwen3technicalreport} as the backbone. For SFT, we used 4$\times$H20 GPUs with a global batch size of 2048. For GRPO, the per-device batch size was set to 8, with a group size $G=16$ (refer to Appendix~\ref{app:params} for full parameters). Notably, we set the KL penalty coefficient $\beta=0$, relying solely on the clipping mechanism for training stability.

\subsection{Offline Performance}


Table~\ref{tab:main} presents the offline performance comparison. While MCA and City-MCA maintain high generation quality (\textsc{QUA}), their recommendation utility (\textsc{HR@12}, \textsc{MRR}) and diversity (\textsc{DIV}) are substantially inferior to generative approaches. City-MCA improves MRR over MCA, confirming that city-specific candidate pools benefit ranking, but it remains constrained by fixed historical candidates. LMP and SPQR further demonstrate the value of location-related signals, yet their reliance on explicit metadata or spatial reformulation limits semantic generalization in local-life query suggestion. LocalSUG consistently outperforms the SFT baseline across all metrics, validating that Beam-Search-Driven GRPO effectively bridges the training-inference gap and enhances multi-objective synergy. Notably, LocalSUG significantly surpasses the original OneSug in both \textsc{HR@12} and \textsc{MRR}. The performance leap of $\text{OneSug}_{rule}$ over OneSug indicates that co-occurrence-based candidates provide stronger dynamic references than RQ-VAE-based strategies in latency-sensitive scenarios. The effects of candidate prompts on performance and the extent to which the LLM creates novel suggestions beyond the provided candidates are further analyzed in Appendix~\ref{sec:cands-beamwidth}. For further performance analysis across different LocalSUG variants, please refer to Appendix~\ref{app:main2}.

\subsection{Ablation Study}

Table~\ref{tab:abl} presents the ablation results on reward components. Most ablations reduce HR, while all of them reduce MRR. Specifically, $r_{rank}$ and $r_{fmt}$ are critical for ranking capability. The absence of $r_{fmt}$ significantly hampers overall performance, likely because the model tends to overfit on high-confidence but malformed low-quality tokens without format constraints. Furthermore, removing the order-aware weight $\omega$ impedes the model's ability to prioritize high-value conversion samples, resulting in suboptimal performance on the \textsc{Order} subset.

\subsection{Online Performance}

\subsubsection{Online A/B Testing}

\begin{table}[!t]
  \centering
  \caption{Online A/B Testing Results of LocalSUG. Significance levels: * p < 0.05, ** p < 0.01, *** p < 0.001.}
  \label{tab:online_ab}
  \resizebox{\columnwidth}{!}{
  \begin{tabular}{llcc}
    \toprule
    \textbf{Category} & \textbf{Metric} & \textbf{Relative Change}\\
    \midrule
    \textbf{User Experience} & No-Result Rate↓ & $-3.98\%^{***}$\\
    \midrule
    \multirow{2}{*}{\textbf{Diversity}} & Unique Item Exposure↑ & $+7.14\%^{***}$\\
    & Category Diversity↑ & $+1.28\%^{***}$ \\
    \midrule
    \multirow{2}{*}{\textbf{Efficiency}} & Session CTR↑ & $+0.27\%^{***}$ \\
    & PV CTR↑ & $+0.35\%^{***}$ \\
    \midrule
    \textbf{User Effort} & Average SUG Input Length↓ & $-0.27\%^{**}$ \\
    \bottomrule
  \end{tabular}
  }
\end{table}

\begin{table*}[!t]
  \centering
  \caption{Online A/B experiment results across business categories.}
  \label{tab:business_category}
  \resizebox{\textwidth}{!}{
  \begin{tabular}{lccccc}
    \toprule
    \textbf{Business Category} & \textbf{PV CTR} & \textbf{Session CTR} & \textbf{Excl.-Orig.} & \textbf{Avg. L2 Cat.} & \textbf{UV CTR} \\
    & & & \textbf{Session CTR} & & \\
    \midrule
    Life Event & +3.15\% & +2.74\% & +3.97\% & +3.50\% & +1.75\% \\
    Beauty \& Medical & +0.32\% & +0.62\% & +0.93\% & +3.74\% & +0.68\% \\
    Transportation & +19.09\% & +15.17\% & +18.93\% & +8.93\% & +13.86\% \\
    Parenting \& Education & +3.81\% & +2.94\% & +3.28\% & +1.86\% & +1.94\% \\
    Entertainment & +0.79\% & +0.65\% & +0.76\% & +3.36\% & +0.38\% \\
    Medicine & +2.69\% & +1.36\% & +1.90\% & +0.91\% & +1.33\% \\
    Merchandise & +3.16\% & +2.43\% & +3.01\% & +1.75\% & +1.65\% \\
    Travel & +1.29\% & +1.06\% & +1.64\% & +2.05\% & +1.27\% \\
    Unknown & -0.90\% & +0.26\% & +0.33\% & +4.42\% & +0.08\% \\
    Movie & +0.63\% & +0.73\% & +1.42\% & +4.00\% & +0.32\% \\
    Hotel \& Homestay & +1.48\% & +0.34\% & +0.15\% & +2.15\% & +1.19\% \\
    Non-transaction & +6.03\% & +4.80\% & +10.50\% & +5.28\% & +3.98\% \\
    Dining & -0.56\% & +0.12\% & +0.30\% & +3.34\% & +0.30\% \\
    \bottomrule
  \end{tabular}
  }
\end{table*}

We deployed LocalSUG in production for a one-month A/B test covering 10\% of real-world traffic. Furthermore, we implemented a fully automated pipeline for daily incremental updates, offline evaluation, and deployment. Requiring minimal manual intervention, this pipeline completes a full update cycle in approximately 9 hours using two H20 GPUs, ensuring high responsiveness to evolving user behaviors.


As shown in Table~\ref{tab:online_ab}, LocalSUG achieves significant improvements. Existing models suffer from data distribution bias and shallow behavioral co-occurrence, which limit diversity and intent capture. LocalSUG mitigates these limitations by expanding the query space and uncovering latent long-tail demand. The online results show that the no-result rate decreases by 3.98\%. System diversity also improves: Unique Item Exposure and Category Diversity increase by 7.14\% and 1.28\%, respectively. By better capturing user intent, LocalSUG improves semantic matching. Consequently, Session CTR and PV CTR increase by 0.27\% and 0.35\%, respectively, while the average SUG input length decreases by 0.27\%. These results demonstrate that users can find relevant results with fewer interactions, thereby improving the overall search experience.

\subsubsection{Performance across Business Categories}
\label{sec:appendix_business_category}

To further evaluate the generalizability of our method, we conduct a fine-grained analysis across different business categories. Table~\ref{tab:business_category} reports the relative improvements over the baseline.

Our method yields positive results across nearly all business categories. Notably, the Transportation and Non-transaction categories exhibit the most substantial improvements, with PV CTR increasing by +19.09\% and +6.03\%, respectively. This indicates that our approach generalizes well across diverse service domains.

\subsubsection{Performance across City Tiers}
\label{sec:appendix_city_tier}

To examine whether our method benefits cities of different development levels, we conduct a comparative analysis across four city tiers (S, A, B, C), where S-tier represents the most developed cities and C-tier the least. Table~\ref{tab:city_tier} reports the relative improvements of our method over the baseline.

\begin{table}[!t]
  \centering
  \caption{Online A/B experiment results across city tiers.}
  \label{tab:city_tier}
  \resizebox{\columnwidth}{!}{
  \begin{tabular}{lcccc}
    \toprule
    \textbf{City Tier} & \textbf{PV CTR} & \textbf{Session CTR} & \textbf{Excl.-Orig.} & \textbf{Avg. L2 Cat.} \\
    & & & \textbf{Session CTR} & \\
    \midrule
    S-tier & +0.01\% & +0.05\% & +0.14\% & +1.35\% \\
    A-tier & 0.00\% & +0.15\% & +0.13\% & +1.27\% \\
    B-tier & +0.35\% & 0.00\% & -0.02\% & +1.29\% \\
    C-tier & +0.63\% & +0.48\% & +0.72\% & +1.60\% \\
    \bottomrule
  \end{tabular}
  }
\end{table}

Our method shows overall positive gains across all city tiers, with stronger advantages in less-developed markets (C-tier). This suggests that the approach is particularly effective in less-developed markets, where query diversity and category coverage yield greater marginal value.

\subsection{Analysis}

\subsubsection{Efficiency of QA-BS and Pruning}
\label{sec:efficiency}

\begin{figure}
    \centering
    
    \begin{subfigure}{0.49\linewidth}
        \centering
        \includegraphics[width=\linewidth]{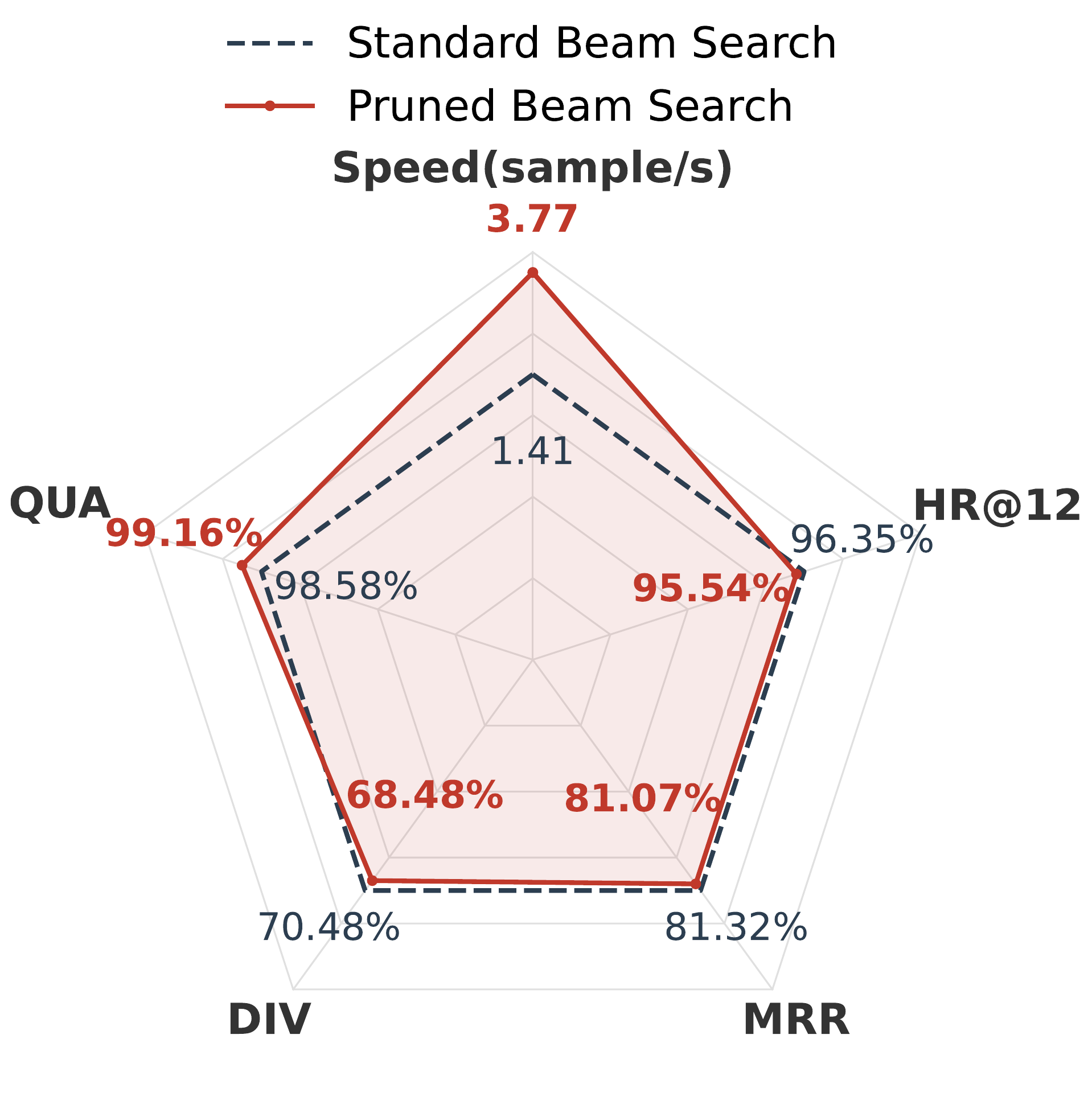}
        \caption{}
    \end{subfigure}
    \hfill
    \begin{subfigure}{0.49\linewidth}
        \centering
        \includegraphics[width=\linewidth]{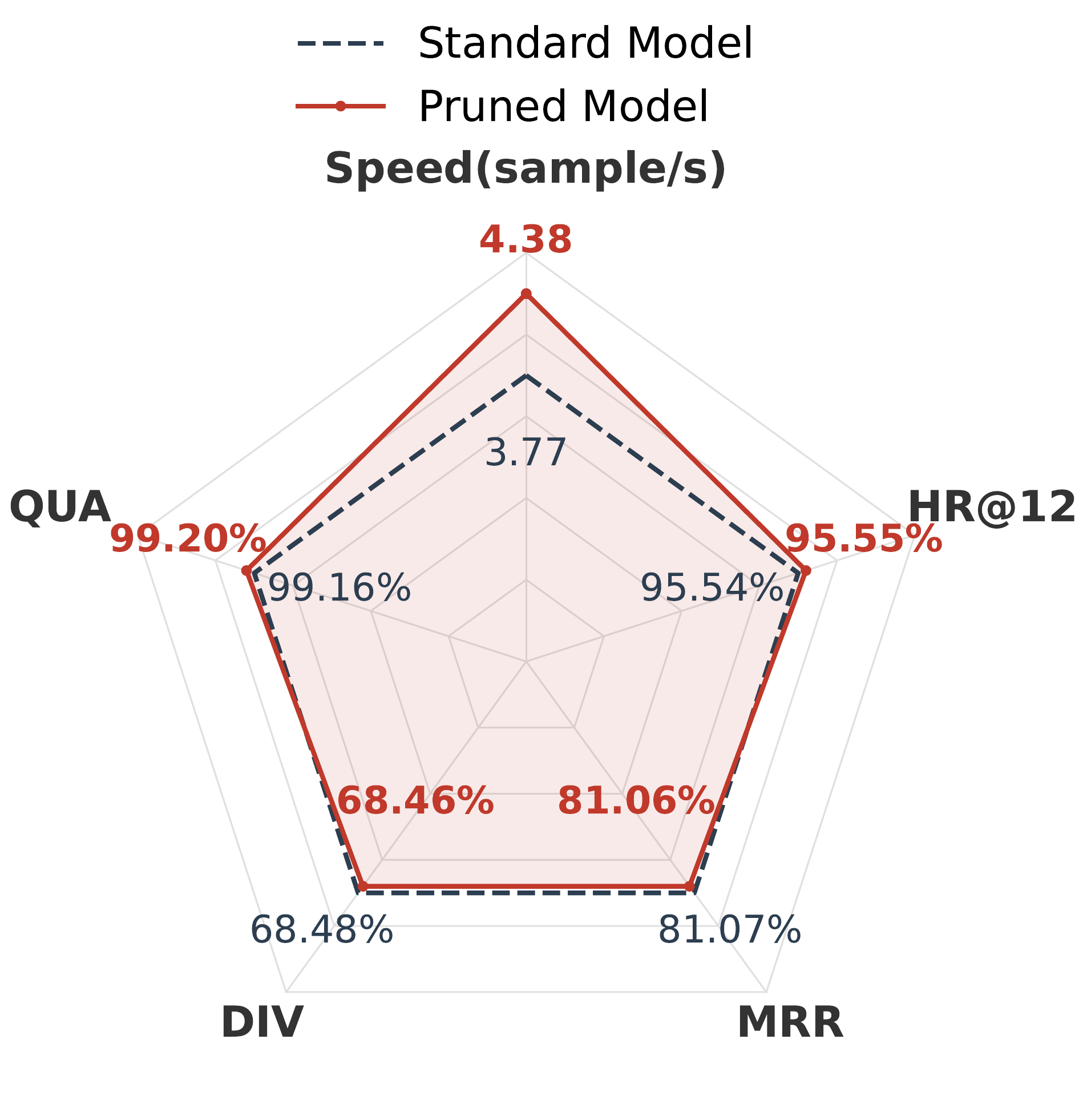}
        \caption{}
    \end{subfigure}
    
    \caption{Impact of (a) QA-BS and (b) vocabulary pruning on inference latency and model performance.}
    \label{fig:bs_ac}
\end{figure}

As shown in Figure~\ref{fig:bs_ac}, QA-BS significantly accelerates inference, increasing the decoding speed from 1.41 to 3.77 samples/s. With vocabulary pruning, the speed is further improved to 4.38 samples/s, resulting in an overall 3.11$\times$ speedup relative to standard beam search. Meanwhile, the acceleration introduces only negligible degradation in HitRate and MRR, with diversity and generation quality remaining stable, demonstrating that the proposed acceleration strategy effectively improves efficiency without compromising model performance.

\subsubsection{Evaluation of User Experience}

\begin{table}[!t]
  \centering
  \caption{Human Ratings of Generated Queries.}
  \label{tab:query_good}
  \resizebox{\columnwidth}{!}{
  \begin{tabular}{lccc}
    \toprule
    
     & Bad & Fair & Good \\
    \midrule
    Control Group & $2.38\%$ & $0.91\%$ & $96.71\%$ \\
    LocalSUG & $1.55\%$ & $0.50\%$ & $97.95\%$ \\
    
    \bottomrule
  \end{tabular}
  }
\end{table}
To strictly measure user satisfaction, we conducted a human evaluation involving three professional annotators located in different cities across the country. The evaluation was performed on an internal annotation platform, where annotators were blind to the source of suggestions (online system or our model) and rated only perceived relevance and diversity. Each sample was independently rated by three annotators, and only unanimous results were retained, yielding a final set of 2,000 samples per method (onlineMCA and LocalSUG). The suggestion queries were rated as Bad, Fair, or Good. As detailed in Table~\ref{tab:query_good}, LocalSUG achieved a higher proportion of 'Good' ratings and a lower rate of 'Bad' and 'Fair' outputs compared to onlineMCA, thereby verifying its effectiveness in enhancing user experience.

\section{Related Work}

Traditional recommendation methods heavily rely on feature interaction design and engineering, optimizing performance through meticulously constructed model architectures~\cite{boldi_query-flow_2008,wide-deep,guo_deepfm_2017,dlrm,din,dien,fan_mobius_2019}. Recent research has introduced generative approaches in specific components such as recall and ranking, breaking through the expressive limitations of traditional models~\cite{tiger,nar4rec,cobra,car,genctr,shi_gensar_2025,chen_pinfm_2025,he_plum_2025,han_mtgr_2025}.

End-to-end generative recommendation is replacing traditional pipelines with unified models. Following the pioneering framework design~\cite{deng_onerec_2025}, recent studies have further extended this paradigm through advanced architectures~\cite{qiu_ega-v1_2025,zheng_ega-v2_2025} and system-level optimizations for better quality and practicality~\cite{zhou_onerec-v2_2025,agarwal_pinrec_2025,li_clear-rec_2025}. Additionally, researchers have adapted generative models to specific tasks, addressing explainability, local services, and semantic integration~\cite{xing_reg4rec_2025,wei_oneloc_2025,car}.

\section{Conclusion}

In this paper, we propose LocalSUG, an end-to-end generative framework for low-latency query suggestion in local-life services. LocalSUG injects term co-occurrence-based city-specific candidates into prompts rather than model parameters, enabling adaptation to evolving city preferences while reducing stale or locally invalid suggestions. To mitigate exposure bias, we introduce Beam-Search-Driven GRPO with multi-objective rewards, and further combine Quality-Aware Accelerated Beam Search with vocabulary pruning for efficient online deployment. Extensive offline evaluations and online A/B testing demonstrate LocalSUG's effectiveness in improving query suggestion, breaking feedback loops, and satisfying long-tail demands in industrial ecosystems.

\section*{Limitations}
Despite its effectiveness, LocalSUG has limitations that motivate future work. First, city-preference-enhanced candidate mining depends on historical logs. Although daily co-occurrence candidates capture evolving preferences, they may not respond quickly to newly opened merchants, temporary closures, or sudden trend shifts. We therefore plan to integrate dense semantic retrieval, real-time merchant status, and external knowledge to address cold-start and rapidly changing scenarios. Second, GRPO relies on manual reward shaping, with coefficients balancing diversity and business metrics. Future work will explore automated or dynamic weight adjustment to achieve a more robust Pareto-optimal balance across different service categories.

\section*{Acknowledgments}
This work was funded by the National Natural Science Foundation of China (NSFC) under Grants No. 62406013 and No. U25B2070, the Beijing Advanced Innovation Center Funds for Future Blockchain and Privacy Computing, and the Fundamental Research Funds for the Central Universities.


\bibliography{custom}

\newpage

\appendix

\section{The Algorithm of Reward Calculation}
Algorithm \ref{alg:reward_pipeline} illustrates the complete reward calculation pipeline.
\label{app:reward_calculation}
\begin{algorithm}[!t]
\caption{Reward Calculation for Beam-Search-Driven GRPO}
\label{alg:reward_pipeline}
\DontPrintSemicolon
\SetKwInOut{Input}{In}\SetKwInOut{Output}{Out}
\Input{Outputs $\mathcal{Y}_G$, Truth $y^*$, Beam $K$, Scores $\mathcal{S}$, Coeffs $\lambda_{\{gap,fmt,rank,hit,miss\}}$}
\Output{Rewards $\mathcal{R} = \{R_1, \dots, R_G\}$}

Initialize $\mathcal{R} \leftarrow \mathbf{0}$, $cnt_{bad} \leftarrow 0$\;

Sort indices $I$ by $\mathcal{S}$ desc; get $rank_i$ for each $y_i$\;
Gap term $v_{gap} = \lambda_{gap} \cdot 1.0$\; 
Tail penalty $v_{tail} = (K \cdot v_{gap}) / (G-K)$\;

\lForEach{$i \in \{1,\dots,G\}$}{
    $R_i \mathrel{+}= v_{\mathrm{gap}}$ if $rank_i \le K$,
    otherwise $R_i \mathrel{-}= v_{\mathrm{tail}}$
}
\ForEach{$i \in \{1 \dots G\}$}{
    \If{$y_i$ is invalid}{
        $R_i \mathrel{-}= \lambda_{fmt}$; 
        
        \lIf{$rank_i \le K$}{$cnt_{bad}\mathrel{++}$}
    }
}

Find the best rank of $y^*$:
$rank^* = \min\{rank_i \mid y_i=y^*\}$, if it exists\;
\uIf{$rank^*$ is not None}{
    idx $\leftarrow$ index of best match\; 
    \textbf{Bonus} $B = \lambda_{rank}/\log_{10}(rank^* + 1)$\;
    \If{$rank^* \le K$}{
        $B \mathrel{+}= \lambda_{hit}$\;
    }
    \Else{
        $B \mathrel{+}= \lambda_{hit}+v_{\mathrm{tail}}$\;
    }    
    $R_{idx} \mathrel{+}= B$\;

    \ForEach{$j$ where $rank_j < rank^*$}{
        $R_j \mathrel{-}=
        \lambda_{\mathrm{rank}}/\log_{10}(rank_j+1)$\;
    }
    
    \ForEach{$j$ where $rank_j > K \land y_j \text{ is valid} \land cnt_{bad} > 0$}{$R_j = \max(R_j, 1)$\; 
    $cnt_{bad}-1$\;}
}
\Else(){
    \ForEach{$i$ where $rank_i \le K/2$}{$R_i \mathrel{=}\min(R_i, -\lambda_{miss})$\;}
    \ForEach{$i$ where $y_i \text{ is valid}$}{$R_i \mathrel{=} \max(R_i, 1)$\;}
}
\Return{$\mathcal{R}$}\;
\end{algorithm}

\section{The Algorithm of Quality-Aware Accelerated Beam Search}
\label{app:beamsearch}
Algorithm \ref{alg:qa_beam_search} illustrates the complete process of the Quality-Aware Accelerated Beam Search.

\begin{algorithm}[!t]
\SetAlgoLined
\DontPrintSemicolon
\caption{Quality-Aware Accelerated Beam Search}
\label{alg:qa_beam_search}
\KwIn{Input $x$, Beam Width $K$, Max Tokens $T$, Absolute Threshold $\tau$, Saturation Coeff $\alpha$, Min Results $R_{min}$, Search Width $K_{search}$, Top-K Window Size $K_{win}$}
\KwOut{Set of generated sequences $Y$}

$B \leftarrow \{ (\text{tokenize}(x), 0.0) \}$\;
$C \leftarrow \emptyset$\;

\For{$t \leftarrow 1$ \KwTo $T$}{
    \lIf{$B$ is empty}{ \textbf{break} }
    
    $Candidates \leftarrow \emptyset$\;
    $C_{temp} \leftarrow \emptyset$ \;
    $S_{win} \leftarrow \text{InitList}(-\infty, \text{size}=K_{win})$ \;
    
    \ForEach{$(seq, score) \in B$}{
        $P(v|seq) \leftarrow \text{Model}(seq)$\;
        \ForEach{token $v \in \text{TopK}(P)$}{
            $s_{new} \leftarrow score + \log P(v|seq)$\;
            $seq_{new} \leftarrow seq + v$\;
            $\text{UpdateSortedWindow}(S_{win}, s_{new})$
            \eIf{$v \in StopIDs$}{
                \If{$s_{new} > \tau$}{
                    Add $(seq_{new}, s_{new})$ to $C_{temp}$\;
                }
            }{
                Add $(seq_{new}, s_{new})$ to $Candidates$\;
            }
        }
    }
    
    \ForEach{$(c, s) \in C_{temp}$}{
        \If{$s > \min(S_{win})$ \textbf{and} $s \ge \tau$}{
            Add $(c, s)$ to $C$\;
        }
    }

    \lIf{$|C| \ge \alpha \cdot K$}{
        \textbf{break}
    }
    
    $S_{max\_cand} \leftarrow \max_{(c,s) \in Candidates} s$\;
    \If{$S_{max\_cand} < \tau$ \textbf{and} $|C| \ge R_{min}$}{
        \textbf{break}\;
    }

    $\mathcal{A} \leftarrow
    \{(c,s)\in \mathit{Candidates} \mid s \ge \tau\}$\;
    $B \leftarrow
    \operatorname{TopK}(\mathcal{A}, K_{\mathrm{search}})$\;
}

$Y \leftarrow \text{Top-}K(C \cup B)$ sorted by score\;
\Return{$Y$}\;
\end{algorithm}

\begin{table}[!t]
    \centering
    \caption{Hyperparameter settings utilized in the proposed framework.}
    \label{tab:hyperparameters}
    \resizebox{0.8\linewidth}{!}{
    \begin{tabular}{l c | l c}
    \toprule
    \textbf{Param} & \textbf{Value} & \textbf{Param} & \textbf{Value} \\
    \midrule
    $m$ & 10 & $n$ & 10\\
    $K$ & 12 & $G$ & 16 \\
    $\epsilon$ & 0.1 & $\delta$ & 1e-4 \\
    $\lambda_{gap}$ & 1.0 & $\lambda_{hit}$ & 1.0 \\
    $\lambda_{rank}$ & 2.0 & $\lambda_{fmt}$ & 4.0 \\
    $\lambda_{miss}$ & 1.0 & $\lambda_{order}$ & 1.5 \\
    $\tau$ & -15 & $\alpha$ & 1.8 \\
    $R_{min}$ & 4 & $K_{win}$ & 15 \\
    $T$ & 15 & $K_{search}$ & 12\\
    $lr$ & 2e-6 & $\beta$ & 0 \\
    
    \bottomrule
    \end{tabular}%
    }
\end{table}

\begin{table*}[!t]
  \centering
  \caption{\label{tab:main2}Offline performance comparison between LocalSUG and various variants. The best results are highlighted in \textbf{bold}, and the second-best results are \underline{underlined}.}
  \resizebox{\textwidth}{!}{
    \begin{tabular}{l|cccc|cccc}           
    
    \toprule
    \multirow{2}{*}{Method} & \multicolumn{4}{c|}{\textsc{Click}} & \multicolumn{4}{c}{\textsc{Order}} \\
     & HR@12↑ & MRR↑ & DIV↑ & QUA↑ & HR@12↑ & MRR↑ & DIV↑ & QUA↑ \\

    \midrule

    $\text{LocalSUG}_{T=0.8}$ & 96.45\% & 80.30\% & 67.34\% & 95.14\% & 97.07\% & 80.24\% & 64.93\% & 94.85\% \\
    $\text{LocalSUG}_{T=1.0}$ & \textbf{96.52\%} & 80.47\% & 68.69\% & 96.96\% & \textbf{97.09\%} & 80.58\% & 66.71\% & 97.25\% \\
    $\text{LocalSUG}_{T=1.2}$ & \underline{96.50\%} & 79.47\% & 68.26\% & 96.68\% & \textbf{97.09\%} & 79.77\% & 66.16\% & 96.73\% \\
    
    $\text{LocalSUG}_{DPO}$ & 95.69\% & \underline{80.97\%} & \textbf{72.83\%} & \underline{97.47\%} & 96.38\% & 81.45\% & \textbf{70.47\%} & \underline{97.74\%} \\
    $\text{LocalSUG}_{Llama}$ & 95.68\% & 80.73\% & 69.90\% & 95.78\% & 96.65\% & \underline{81.65\%} & 68.07\% & 96.43\% \\
    LocalSUG & 96.36\% & \textbf{81.13\%} & \underline{71.49\%} & \textbf{98.55\%} & \textbf{97.09\%} & \textbf{81.80\%} & \underline{69.31\%} & \textbf{98.78\%} \\
                
    \bottomrule
    \end{tabular}%
    }

\end{table*}%

\section{Metric Definitions}
\label{sec:appendix_metrics}

We formally define the evaluation metrics used in our experiments. Let $\mathcal{T}$ be the test set of size $N$. For each input $i$, the model generates a ranked list of $K$ queries, denoted as $\mathcal{Q}_i = [q_{i,1}, q_{i,2}, \dots, q_{i,K}]$.

\paragraph{HitRate@K (HR@K) and MRR}
HR@K calculates the percentage of cases where the ground truth $Y_i$ is present in the generated list $\mathcal{Q}_i$. MRR computes the mean reciprocal rank of the first correct answer.
\begin{align}
    \text{HR}@K &= \frac{1}{N} \sum_{i=1}^{N} \mathbb{I}(Y_i \in \mathcal{Q}_i), \\
    \text{MRR}
        &= \frac{1}{N}\sum_{i=1}^{N}
        \frac{\mathbb{I}(Y_i \in \mathcal{Q}_i)}{\operatorname{rank}_i},
\end{align}
where $\text{rank}_i$ denotes the position of $Y_i$ in $\mathcal{Q}_i$.

\paragraph{Diversity (DIV)}
DIV measures the richness of the generated vocabulary. It is the ratio of unique query strings across the entire dataset to the total number of generated queries:
\begin{equation}
    \text{DIV} = \frac{|\text{Unique}(\bigcup_{i=1}^{N} \mathcal{Q}_i)|}{N \times K}.
\end{equation}

\paragraph{Quality (QUA)}
This metric evaluates the validity of individual generated queries at the query level, aligning with the \textit{Format Penalty} defined in our GRPO objective. We first define a normalization function $\text{norm}(\cdot)$ (e.g., lowercasing and removing punctuation). Let $\tilde{q}_{i,j} = \text{norm}(q_{i,j})$.

A generated query $q_{i,j}$ is considered effective only if it satisfies two conditions:
(1) $\mathcal{F}(q_{i,j})$: The raw query is free from structural errors (e.g., \textbf{invalid tokens}, garbled text).
(2) $\mathcal{U}(q_{i,j})$: The normalized query is unique relative to higher-ranked queries in the same list (i.e., $\tilde{q}_{i,j} \notin \{\tilde{q}_{i,1}, \dots, \tilde{q}_{i,j-1}\}$).

The metric is calculated as:
\begin{equation}
\begin{split}
    \text{QUA} = \frac{1}{N \times K} \sum_{i=1}^{N} \sum_{j=1}^{K} \mathbb{I}\Big( & \mathcal{F}(q_{i,j}) \land \\
    & \tilde{q}_{i,j} \notin \{\tilde{q}_{i,m}\}_{m<j} \Big).
\end{split}
\end{equation}
This formulation ensures that intra-list duplicates and malformed outputs are strictly penalized.

\paragraph{Online Metrics}

\begin{itemize}
    \item \textbf{No-Result Rate:} The frequency of requests that yield no results.
    \item \textbf{Unique Item Exposure:} The average number of de-duplicated exposed items per user, measuring the breadth of item discovery.
    \item \textbf{Category Diversity:} The average number of distinct L2 categories exposed per user, capturing the diversity of recommendation coverage.
    \item \textbf{PV CTR:} The ratio of total clicks to total recommendation impressions across all page views.
    \item \textbf{Session CTR:} The proportion of sessions containing at least one click, reflecting overall user engagement at the session level.
    \item \textbf{Average SUG Input Length:} The average input character length per user.
\end{itemize}

\section{Baseline Details}
\label{app:baseline}

We compare LocalSUG with the following baselines.

\paragraph{MCA}
MCA is the standard industrial multi-stage cascading architecture. It follows a recall-ranking pipeline, where BGE\footnote{https://huggingface.co/BAAI/bge-base-zh-v1.5}~\cite{bge} is used for candidate retrieval and DIN~\cite{din} is used for ranking.

\paragraph{City-MCA}
City-MCA is a city-enhanced variant of MCA. Compared with MCA, it uses city-specific candidate pools during recall and ranking, allowing the traditional cascading system to exploit city-level preference signals.

\paragraph{LMP}
LMP is adapted from location metadata-based personalization in web search~\cite{lmp}. It incorporates user location metadata as an additional personalization signal for query suggestion, enabling location-conditioned ranking over candidate queries.

\paragraph{SPQR}
SPQR is adapted from spatial proximity query reformulation~\cite{spqr}. It uses location-aware reformulation signals to enhance query suggestions, emphasizing spatially relevant query alternatives.

\paragraph{SFT}
SFT denotes the backbone model fine-tuned with standard supervised learning on query suggestion logs, without preference optimization.

\paragraph{OneSug}
OneSug is a reproduction of the state-of-the-art end-to-end generative query suggestion framework~\cite{guo_onesug_2025}. It directly generates ranked suggestion lists with a unified generative model.

\paragraph{OneSug$_{rule}$}
OneSug$_{rule}$ replaces the original OneSug candidate construction strategy with our co-occurrence rule-based candidates, allowing us to isolate the effect of candidate quality.

\paragraph{LocalSUG Variants}
We further evaluate several LocalSUG variants. LocalSUG$_{DPO}$ replaces Beam-Search-Driven GRPO with DPO~\cite{dpo}. LocalSUG$_{Llama}$ replaces Qwen3-0.6B with Llama-3.2-1B-Instruct\footnote{https://huggingface.co/meta-llama/Llama-3.2-1B-Instruct}. We also test different sampling strategies during GRPO exploration while keeping deterministic beam search for inference.

\section{Hyperparameter Settings}
\label{app:params}

Table \ref{tab:hyperparameters} presents the key hyperparameters used for both training and inference.

\section{Comparison with LocalSUG variants}
\label{app:main2}

    


    
                
    
As shown in Table~\ref{tab:main2}, we compare LocalSUG with several variants. The observation that substituting beam search with random sampling substantially diminishes MRR and diversity underlines the advantages of training-inference sampling consistency. Moreover, we find that vanilla DPO training results in a clear performance drop in Hit Rate and MRR. Notably, our framework maintains competitive performance with a Llama backbone, highlighting its generalizability across various large language models.

\section{Additional Analysis}

\subsection{Impact of Context and Decoding Width}
\label{sec:cands-beamwidth}

\begin{figure}[!t]
\label{fig:cands_beamwidth}
    \centering
    \begin{subfigure}{0.9\linewidth}
        \centering
        \includegraphics[width=\linewidth]{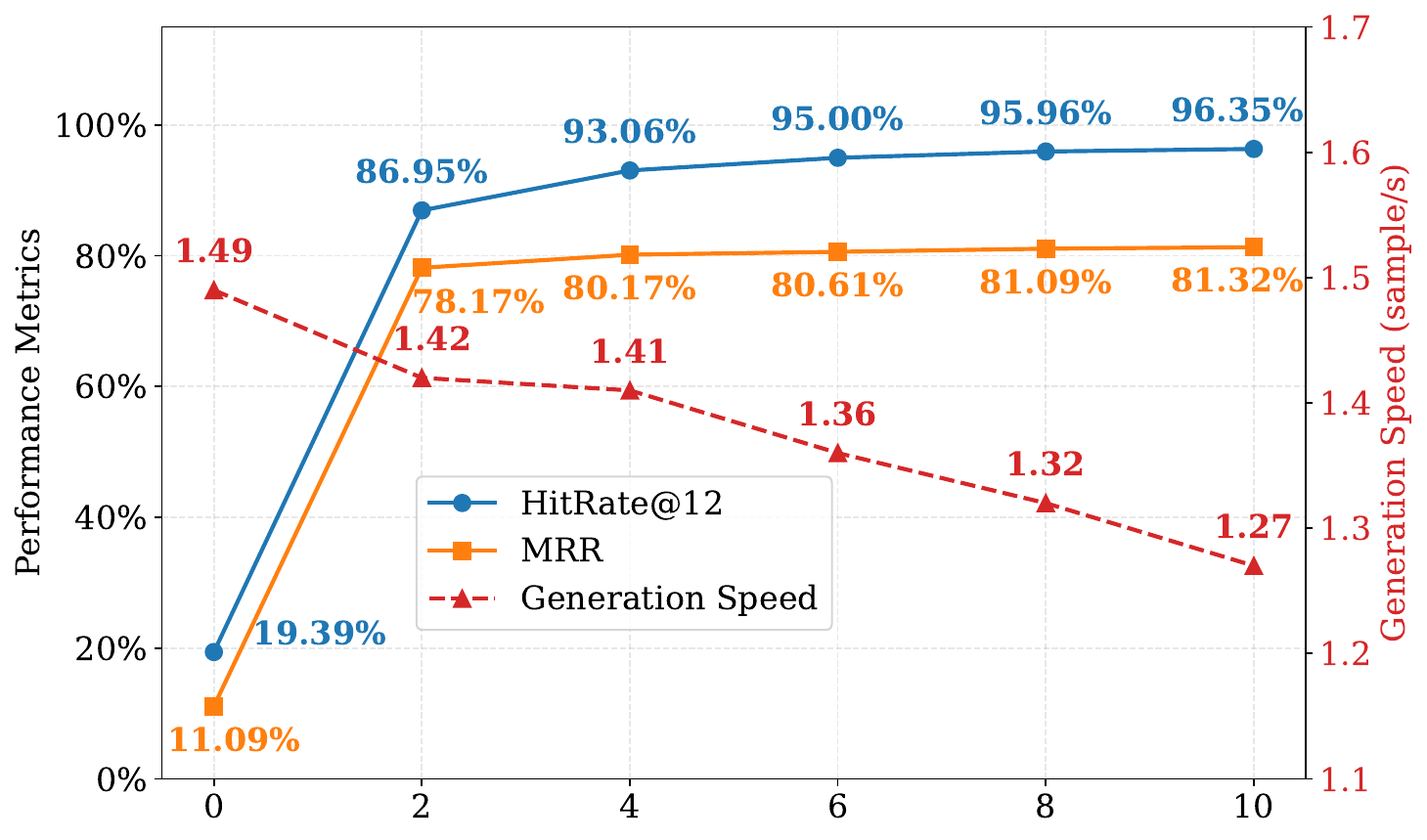}
        \caption{Candidate Count $|\mathcal{C}_{cands}|$}
        \label{fig:cands}
    \end{subfigure}   
    \hfill
    \begin{subfigure}{0.9\linewidth}
        \centering
        \includegraphics[width=\linewidth]{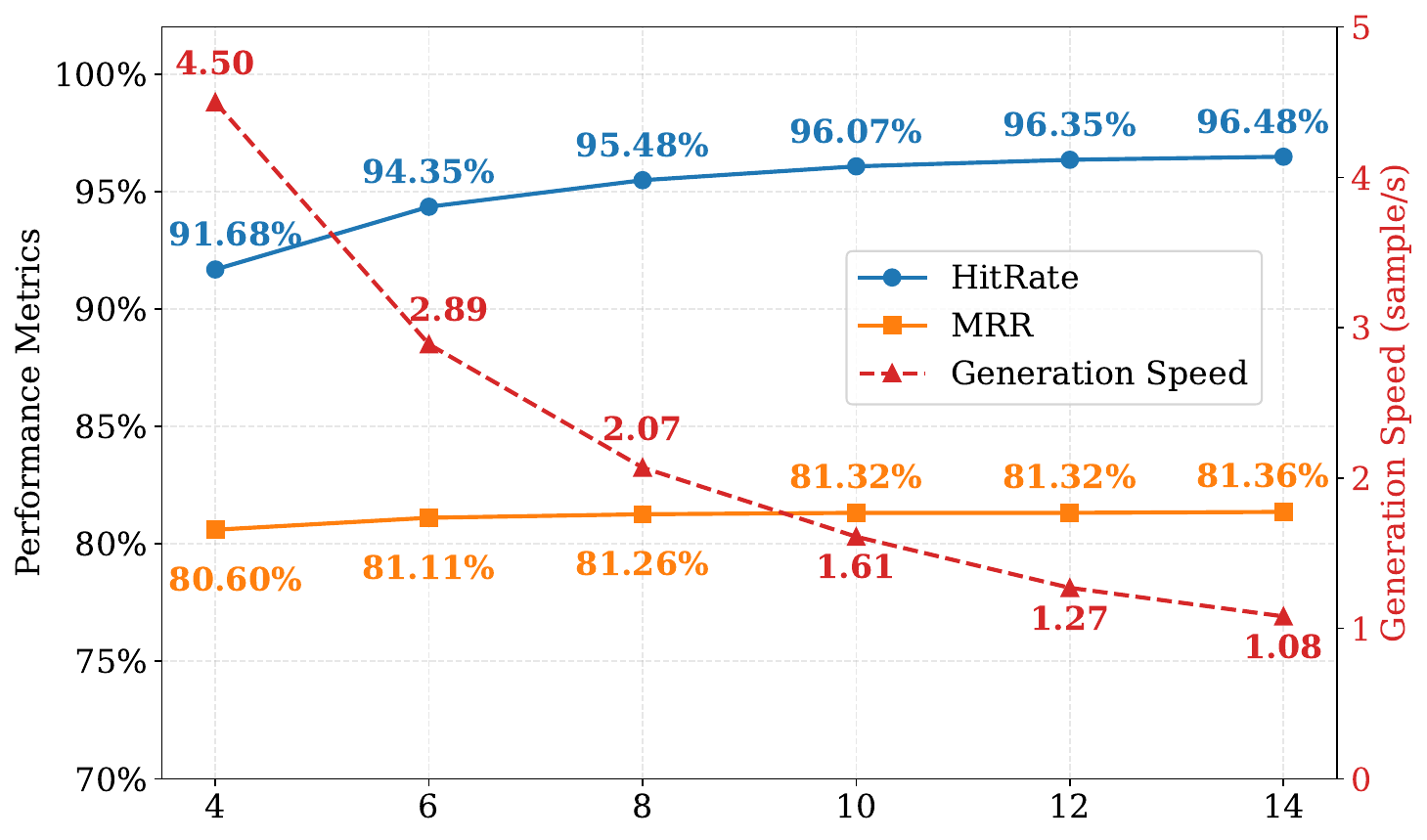}
        \caption{Beam Width}
        \label{fig:beamwidth}
    \end{subfigure}
    \caption{Sensitivity analysis of retrieval candidates and beam search width on efficiency and effectiveness.}
\end{figure}

We analyze the sensitivity of the model to candidate count ($m$) and beam width ($K$) using the \textsc{Mix} dataset.
As shown in Figure~\ref{fig:cands}, the number of candidates has a minimal impact on inference speed, as input length is dominated by user history. While the model collapses without candidates, providing merely $m=2$ candidates remarkably recovers 90.2\% of the optimal HitRate and 96.1\% of MRR, demonstrating robustness to scenarios with limited candidate availability.
We further analyze the source of generated suggestions. We find that 41\% of the generated queries do not exactly appear in the provided candidate set, indicating that the model does not simply copy all prompt candidates as valid outputs. Instead, it selectively uses candidates as city-preference references and further composes new suggestions based on user-side context. For example, if a user frequently consumes rice noodles and has recently visited MixC, the model may generate a new query such as ``MixC rice noodles'', which is absent from the candidate set but consistent with both the user's historical preference and recent behavior. In our logs, such a generated query was eventually clicked by the user, demonstrating LocalSUG's ability to create personalized and context-aware suggestions beyond candidate copying.
Figure~\ref{fig:beamwidth} highlights the latency-performance Pareto frontier. Reducing $K$ significantly boosts speed with marginal performance loss; specifically, $K=6$ retains 97.9\% of the performance of $K=12$ while achieving a 2.28$\times$ speedup.


\begin{table}[!t]
  \centering
  \caption{Ablation results (mean $\pm$ std) over 3 random seeds on the \textsc{Mix} dataset.}
  \label{tab:ablation_std}
  \resizebox{\columnwidth}{!}{
  \begin{tabular}{l|cc}
    \toprule
    \textbf{Model} & \textbf{HR@12} & \textbf{MRR} \\
    \midrule
    LocalSUG & 96.12\%$\pm$0.03\% & 81.40\%$\pm$0.02\% \\
    w/o $r_{gap}$ & 95.54\%$\pm$0.04\% & 81.21\%$\pm$0.11\% \\
    w/o $r_{hit}$ & 95.79\%$\pm$0.11\% & 81.30\%$\pm$0.03\% \\
    w/o $r_{rank}$ & 96.53\%$\pm$0.00\% & 80.13\%$\pm$0.11\% \\
    w/o $r_{fmt}$ & 95.83\%$\pm$0.03\% & 76.20\%$\pm$0.83\% \\
    w/o $r_{miss}$ & 95.90\%$\pm$0.02\% & 81.36\%$\pm$0.01\% \\
    w/o $\omega$ & 96.14\%$\pm$0.03\% & 81.29\%$\pm$0.08\% \\
    \midrule
    \textbf{Model} & \textbf{DIV} & \textbf{QUA} \\
    \midrule
    LocalSUG & 72.05\%$\pm$0.04\% & 98.30\%$\pm$0.06\% \\
    w/o $r_{gap}$ & 72.31\%$\pm$0.02\% & 97.66\%$\pm$0.05\% \\
    w/o $r_{hit}$ & 71.65\%$\pm$0.31\% & 97.86\%$\pm$0.27\% \\
    w/o $r_{rank}$ & 70.00\%$\pm$0.07\% & 99.00\%$\pm$0.02\% \\
    w/o $r_{fmt}$ & 62.47\%$\pm$0.18\% & 89.04\%$\pm$0.09\% \\
    w/o $r_{miss}$ & 71.58\%$\pm$0.02\% & 97.93\%$\pm$0.03\% \\
    w/o $\omega$ & 71.72\%$\pm$0.50\% & 97.76\%$\pm$0.47\% \\
    \bottomrule
  \end{tabular}
  }
\end{table}

\subsection{Robustness to Group Size \texorpdfstring{$G$}{G}}

\begin{figure}[!t]
    \centering
    \includegraphics[width=\linewidth]{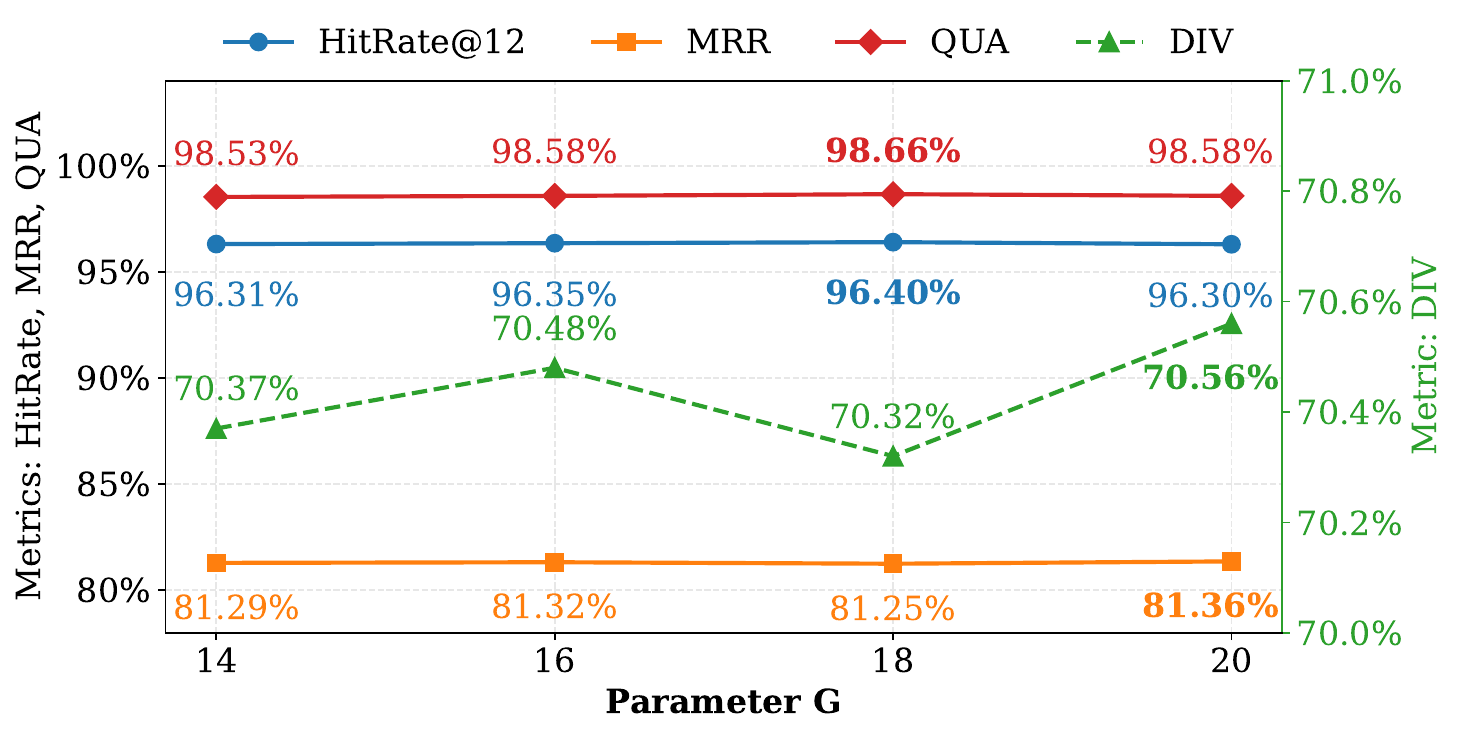}
    \caption{Performance stability across different GRPO group sizes ($G$).}
    \label{fig:g}
\end{figure}

In our Beam-Search-Driven GRPO, the group size $G$ determines the ratio of negative samples during training. We investigated the impact of varying $G$ from 14 to 20 on the \textsc{Mix} dataset. As illustrated in Figure~\ref{fig:g}, aside from minor fluctuations in Diversity (DIV), performance across all key metrics remains highly stable. This indicates that our method is robust to the hyperparameter $G$, maintaining effectiveness without requiring extensive tuning of the group size.

\subsection{Ablation with Multiple Random Seeds}
\label{sec:ablation_std}

To address the concern that minor metric differences in our original ablation study (Table~\ref{tab:abl}) might be attributed to training noise, we re-conducted all ablation experiments using three different random seeds and report the mean and standard deviation on the \textsc{Mix} dataset. As shown in Table~\ref{tab:ablation_std}, the consistently low standard deviations across all variants indicate stable training, ruling out randomness as a factor. The results reaffirm that the format and rank rewards are the most critical components: removing the format reward (\textit{w/o $r_{fmt}$}) leads to severe degradation in MRR ($-5.20\%$) and QUA ($-9.26\%$), while removing the rank reward (\textit{w/o $r_{rank}$}) causes the largest MRR drop ($-1.27\%$).

\end{document}